\documentclass[12pt]{article}
\usepackage{amsmath,amsfonts}
\usepackage{algorithmic}
\usepackage{algorithm}
\usepackage{array}
\usepackage[caption=false,font=normalsize,labelfont=sf,textfont=sf]{subfig}
\usepackage{textcomp}
\usepackage{stfloats}
\usepackage{url}
\usepackage[normalem]{ulem}
\usepackage{scalerel}
\usepackage{graphicx}
\usepackage{verbatim}
\usepackage{xcolor}
\usepackage{lettrine}
\usepackage{float}
\usepackage{cite}
\footnotesize
\usepackage{authblk} 
\usepackage{cite}
\usepackage{amsmath,amssymb,amsfonts}
\usepackage{algorithm}
\usepackage{algorithmic}
\usepackage{graphicx}
\usepackage{multirow}
\usepackage{booktabs}
\usepackage{textcomp}
\usepackage{booktabs, multirow, xcolor, colortbl, array, rotating}
\usepackage[colorlinks=true, citecolor=blue, linkcolor=blue, urlcolor=blue]{hyperref}

\begin{document}

\title{SpatialMAGIC: A Hybrid Framework Integrating Graph Diffusion and Spatial Attention for Spatial Transcriptomics Imputation}

\author[1]{Sayeem Bin Zaman}
\author[1]{Fahim Hafiz}
\author[1]{Riasat Azim\thanks{Corresponding author: \href{mailto:riasat@cse.uiu.ac.bd}{riasat@cse.uiu.ac.bd}}}

\affil[1]{Department of Computer Science and Engineering, United International University, United City, Madani Avenue, Badda, Dhaka 1212, Bangladesh}

\date{} 
\maketitle




\newcommand{\mymethod}{SpatialMAGIC}
\newcommand{\stereoseq}{Stereo-seq}
\newcommand{\slideseq}{Slide-seq}
\newcommand{\scispace}{Sci-space}

\begin{abstract}

Spatial transcriptomics (ST) enables mapping gene expression with spatial context but is severely affected by high sparsity and technical noise, which conceals true biological signals and hinders downstream analyses. To address these challenges, \mymethod{} was proposed, which is a hybrid imputation model combining MAGIC-based graph diffusion with transformer-based spatial self-attention. The long-range dependencies in the gene expression are captured by graph diffusion, and local neighborhood structure is captured by spatial attention models, which allow for recovering the missing expression values, retaining spatial consistency. Across multiple platforms, \mymethod{} consistently outperforms existing baselines, including MAGIC and attention-based models, achieving peak Adjusted Rand Index (ARI) scores in clustering accuracy of 0.3301 on high-resolution \stereoseq{} data, 0.3074 on \slideseq{}, and 0.4216 on the \scispace{} dataset. Beyond quantitative improvements, \mymethod{} substantially enhances downstream biological analyses by improving the detection of both up- and down-regulated genes while maintaining regulatory consistency across datasets. The pathway enrichment analysis of the recovered genes indicates that they are involved in consistent processes across key metabolic, transport, and neural signaling pathways, suggesting that the framework improves data quality while preserving biological interpretability. Overall, \mymethod{}'s hybrid diffusion attention strategy and refinement module outperform state-of-the-art baselines on quantitative metrics and provide a better understanding of the imputed data by preserving tissue architecture and uncovering biologically relevant genes. The source code and datasets are provided in the following link: \url{https://github.com/sayeemzzaman/SpatialMAGIC}

\end{abstract}


\section{Introduction}
Spatial transcriptomics (ST) is an advanced sequencing technology that enables the mapping of gene expression within its maintained spatial context, providing deep insights into tissue heterogeneity, cell–cell interactions, and disease mechanisms. Despite its potential, high-resolution ST data, such as that produced by Stereo-seq, is severely impacted by noise and extreme sparsity, with studies indicating that more than 84\% of gene expression values can be zero \cite{luo2025stinr,xu2024unsupervised}. Effective imputation methods are therefore essential to reveal biologically significant signals and facilitate accurate downstream analyses, including spatial clustering, trajectory inference, and tissue domain detection.

Various computational strategies have been proposed to mitigate these challenges. Initial efforts often adapted single-cell RNA sequencing (scRNA-seq) methods, such as MAGIC, which employs Markov affinity-based graph diffusion \cite{dijk2017magic}. More recently, graph-based and deep learning models like SEDR (deep autoencoder and variational graph autoencoder) and STINR (implicit neural representation) have shown promise in capturing spatial relations \cite{xu2024unsupervised,luo2025stinr}. Other approaches, such as DiffusionST and SpotDiff, utilizes generative diffusion models to enhance data quality, though they often face significant computational costs and scalability issues when applied to large-scale datasets exceeding 50,000 spots. A critical trade-off persists in the field and many models either struggle to represent complex, multi-modal relationships or fail to balance global diffusion with local structural preservation.

To bridge this gap, we developed \mymethod{}, a customised imputation pipeline that simultaneously utilises gene expression patterns and spatial coordinates. We developed a customized imputation pipeline that integrates MAGIC, a Markov affinity-based imputation method, with a spatial attention-guided neural network to utilize both gene expression patterns and spatial data simultaneously. The proposed method natively assists the extreme sparsity and spatial heterogeneity common in \stereoseq{}, \slideseq{}, and \scispace{} dataset, improving clustering accuracy and enabling more refined resolution of tissue structure. The key contributions of this work are summarized as follows:

\begin{itemize}
    \item We introduce \textbf{\mymethod{}}, a hybrid imputation framework that integrates MAGIC-based graph diffusion with transformer-driven spatial self-attention combinedly. This design enables the simultaneous modeling of local microenvironments and long-range tissue dependencies, effectively addressing the inherent sparsity and noise in ST data.
    
    \item We develop a novel spatial fusion refinement module that unifies expression and spatial embeddings through an encoder–decoder architecture. This module refines the initial imputations to reconstruct biologically coherent expression landscapes by learning nonlinear mappings that transform the integrated features into a unified latent representation.
    
    \item We present an extensive multi-platform and multi-resolution evaluation across three major ST technologies, \stereoseq{}, \slideseq{}, and \scispace{}, under diverse biological conditions. \mymethod{} outperforms standard graph-based imputation and hybrid attention-based baselines (including MAGIC, Attention PCA, and Attention UMAP) in terms of denoising quality, clustering accuracy, and spatial domain recovery.
    
    \item We demonstrate that \textbf{\mymethod{}} substantially improves downstream spatial analyses, including the identification of differentially expressed genes (DEGs) and fine-grained tissue domain representation, thereby enabling deeper insights into cellular organization and tissue architecture.
    
    \item Beyond imputation, we provide an interpretable framework that leverages spatial attention maps and graph diffusion patterns to uncover biologically meaningful long-range interactions, offering a new perspective on tissue-level gene regulation.
\end{itemize}

\noindent The remainder of this paper is organized as follows: Section~\ref{sec2} reviews related work on spatial transcriptomics imputation, graph-based methods, and transformer-based approaches. Section~\ref{sec3} describes the proposed \textbf{\mymethod{}} framework, including dataset descriptions, preprocessing steps, MAGIC-based graph diffusion, spatial attention embedding, and the fusion-based refinement module. Section~\ref{sec4} reports the results, including quantitative performance comparisons and qualitative analyses of imputed spatial patterns. Finally, Section~\ref{sec5} discusses the implications, limitations, and potential extensions of the proposed method.

\section{Literature Review}\label{sec2}

The advancement of spatial transcriptomics has enabled the high-resolution mapping of gene expression within tissues, yet the field is persistently challenged by high dropout rates and technical noise, which results in sparse datasets \cite{hu2023adept, chen2025spotdiff, cui2025diffusionst, duan2024impeller, lu2025stgrl, song2023gntd}. Prior computational efforts to address these problems have mostly borrowed computation methods used in single-cell RNA sequencing (scRNA-seq), like MAGIC, which uses Markov affinity-based graph diffusion to exchange information among similar cells \cite{duan2024impeller, cui2025diffusionst, dijk2017magic}. Although capable of capturing transcriptional signatures, such approaches often miss the detailed spatial information needed to model cell-to-cell interactions \cite{duan2024impeller}.

In order to utilize spatial information, different graph-based and deep learning models have been developed throughout the time. ADEPT makes use of a graph autoencoder (GAE) that applies iterative clustering of differentially expressed genes (DEGs) to reduce variance and imputation \cite{hu2023adept}. ADEPT performed better on any dorsal lateral prefrontal cortex (DLPFC) and breast cancer data and did better than other tools, such as SEDR and STAGATE in terms of Adjusted Rand Index (ARI) measures. But it is not scalable, meaning that the higher the number of genes and clustering, the higher its run time. In the same way, Impeller constructs a heterogeneous graph which reflects spatial similarity plus expression similarity, with a learnable path operator to prevent the over-smoothing problem of legacy Laplacian matrices \cite{duan2024impeller}. Impeller proved to be more accurate on L1 distance and RMSE in terms of imputation than gene expression only algorithms, such as eKNN.

Recently, deep generative and diffusion-based models have become effective alternatives to ST enhancement. DiffusionST combines a graph convolutional network (GCN) and a zero-inflated negative binomial (ZINB) distribution to cleanse data, and a diffusion model to optimize expression profiles \cite{cui2025diffusionst}. It was able to handle clustering with high accuracy (ARI scores of 0.43 to 0.65) and it was well resistant to dropout noise that had been introduced manually. However, it has a higher computational cost and thus it is not very appropriate when working with large-scale datasets (over 50,000 spots). SpotDiff suggests using a multi-modal conditional diffusion model with a spot-gene prompt learning module to learn associations between spots and genes with incorporating scRNA-seq data \cite{chen2025spotdiff}. It always performed better in Pearson Correlation Coefficient (PCC) and Structural Similarity Index (SSIM) than Tangram and gimVi. In another method, SpaIM, the style transfer learning method is used to isolate shared material and modality-specific style and attains an overall accuracy score of 0.95 on 53 datasets \cite{li2025spaim}. Nonetheless, it can still be extended to more complex architectures such as graph transformers since it is based on simple multi-layer perceptron (MLP) layers.

Other models are based on multi-task optimization and tensor models, for example, stGRL uses multi-task graph contrastive representation learning to combine domain identification, denoising, and imputation \cite{lu2025stgrl}. It has the median Normalized Mutual Information (NMI) score o f0.69 on benchmark datasets, but its hardware consumption of GPU memory is a limiting factor. GNTD is a neural-network-based method that employs graph-guided neural-tensor decomposition, and it models the data as a three-way Tensor regularised by the spatial and protein-protein interaction (PPI) networks \cite{song2023gntd}. Although GNTD has continued to provide higher improvements in cross-validations, it needs to be placed in a grid format, which restricts its use to non-grid systems. Another architecture, SpateCV is based on a conditional variational autoencoder (CVAE) with alignment regularization to co-embed scRNA-seq and ST data \cite{yuan2025spatecv}. Although it is the best able to reconstruct spatial patterns, the quality of its result strongly relies on the quality of the reference scRNA-seq data.

Altogether, the current ST imputation systems have already achieved much in using graph neural networks, generative adversarial networks, and diffusion models. There is however a critical trade-off between computational efficiency and the capability to represent complex, multi-modal relationships. Many models either struggle with scalability on high-resolution datasets, rely excessively on matched reference data, or fail to balance global diffusion with local structural preservation. \mymethod{} uses these gaps to combine MAGIC-based graph diffusion with transformer-based spatial attention with a molecular-spatial fusion module, enabling high-fidelity imputation that preserves tissue architecture and biological interpretability.

\section{Methodology}\label{sec3}
The Methodology section addresses the overall framework of \mymethod{}. The proposed method utilizes spatial information and gene expression for appropriate imputation by integrating transformer-based attention mechanisms along with diffusion approach in a previous method, MAGIC \cite{dijk2017magic}. We demonstrate that this approach achieves improved resolution of the spatial domain and provides biological signal recovery. We first discuss the ST datasets and the respective pre-processing strategies. Then, we discuss the details of our proposed method, \mymethod{}. 

\begin{figure*}[ht!]
  \centering
  \includegraphics[width=1\textwidth,height=.5\textheight]{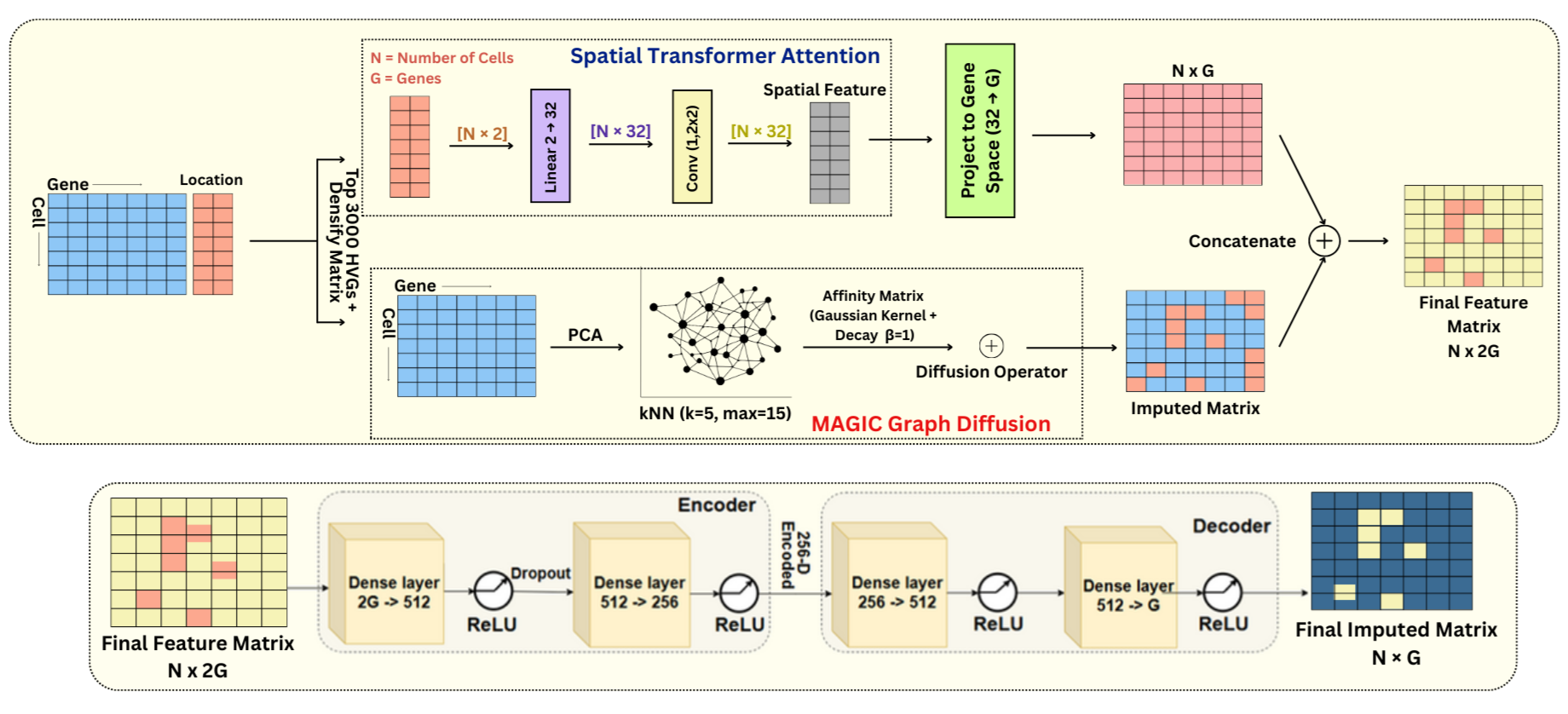}
  \caption{Overview of the \mymethod{} framework. The input gene expression matrix undergoes MAGIC-based diffusion to recover local gene patterns. Simultaneously, spatial coordinates are embedded using a transformer encoder to capture spatial relationships. The outputs from both branches are fused and passed through a neural decoder to reconstruct an enhanced expression profile for downstream spatial clustering and analysis.}
 \label{fig: pipeline}
\end{figure*}


\subsection{Datasets}

To evaluate the performance and generalizability of \textbf{\mymethod{}}, we utilized a diverse set of publicly available ST datasets (Table \ref{tab:datasets}) spanning different sequencing technologies and biological conditions. These include high-resolution developmental and adult tissue datasets from the \textbf{\stereoseq{}} \cite{stereoseq2_xu2024spatiotemporal} platform, disease and control samples from the \textbf{\slideseq{}} \cite{slideseq3_stickels2021highly} platform, and a single-cell spatially resolved dataset from the \textbf{\scispace{}} \cite{scispace4_srivatsan2021embryo} platform. They are selected due to their wide level of use in previous computational work as well as the availability of their annotation data, which is crucial for evaluating unsupervised learning/clustering methods. Together, these datasets encompass a wide range of biological complexity, spatial resolutions, and tissue types, providing a robust benchmark for testing imputation accuracy, clustering performance, and spatial domain recovery. The details of the datasets are described below as well as in Table \ref{tab:datasets}:

\begin{table*}[ht!]
\caption{Summary of ST Datasets Used for Experiment}
\label{tab:datasets}
\centering
\resizebox{\textwidth}{!}{
\begin{tabular}{lllc c}
\toprule
\textbf{Platform} & \textbf{Dataset Name} & \textbf{Biological Description} & \textbf{Cells} & \textbf{Genes} \\
\midrule
\multirow{3}{*}{\stereoseq{} \cite{stereoseq2_xu2024spatiotemporal}} 
  & DT2\_D0 & Embryonic Day 9.5 & 42{,}658 & 20{,}735 \\
  & DX6\_D2 & Embryonic Day 11.5 & 14{,}852 & 19{,}430 \\
  & FB2\_D1 & Adult mouse brain tissue & 16{,}263 & 19{,}639 \\
\midrule
\multirow{3}{*}{\slideseq{} \cite{slideseq3_stickels2021highly}} 
  & slide\_seq\_diabetes1\_T4 & Diabetic mouse model (T1D) & 9{,}435 & 16{,}884 \\
  & slide\_seq\_WT1\_T3 & Healthy wild-type control & 9{,}008 & 16{,}598 \\
  & stickles\_mouse\_slideseq & V2 mouse brain benchmark & 22{,}650 & 3{,}919 \\
\midrule
\scispace{} \cite{scispace4_srivatsan2021embryo} & GSE166692\_scispace & Embryonic mouse tissue & 9{,}517 & 24{,}879 \\
\bottomrule
\end{tabular}
}
\end{table*}

\begin{itemize}
    \item \textbf{\stereoseq{} Datasets}:  
    Three samples were selected from mouse embryonic and adult tissues from \stereoseq{} technology \cite{stereoseq2_xu2024spatiotemporal}. DT2\_D0 corresponds to embryonic day 9.5 (E9.5), representing an early stage of organogenesis. DX6\_D2, collected at approximately E11.5, captures a mid-developmental stage with advanced tissue differentiation. FB2\_D1 is a mature brain section, serving as a reference for fully developed tissue organization.

    \item \textbf{\slideseq{} Datasets}:  
    We included three datasets from the \slideseq{} and Slide-seqV2 platforms \cite{slideseq3_stickels2021highly}. The \textit{slide\_seq\_diabetes1\_T4} sample originates from a type 1 diabetes mouse model, while \textit{slide\_seq\_WT1\_T3} serves as the healthy control. \textit{stickles\_mouse\_slideseq} comes from the seminal Slide-seqV2 study and serves as a high-resolution benchmark of adult mouse tissue, typically covering brain regions.

    \item \textbf{\scispace{} Dataset}:  
    The \textit{GSE166692\_scispace} dataset comprises spatially resolved single-cell data from over 120,000 nuclei in mouse embryos. Generated using \scispace{} technology \cite{scispace4_srivatsan2021embryo}, it enables fine-grained analysis of early developmental processes and spatial tissue patterning at single-cell resolution.
\end{itemize}

\subsection{Data Preprocessing}

The gene expression matrix in each dataset is represeted by $\mathbf{X} \in \mathbb{R}^{n \times g}$, where $n$ is the number of spots and $g$ is the total genes. The preprocessing steps include library size normalization to correct for sequencing depth differences and log-transformation to stabilize variance. After these steps, dimensionality was further reduced by selecting the top $k = 3000$ highly variable genes to remove low-variability genes that contribute limited information. Gene variability was computed across all spots, and genes were ranked accordingly. The highest-ranked genes were retained, yielding the filtered expression matrix $\mathbf{X}_{\mathrm{HVG}} \in \mathbb{R}^{n \times k}$, where $k \ll g$.

Since $\mathbf{X}_{\mathrm{HVG}}$ contains a high percentage of zero values resulting from technical dropouts, the matrix was converted into dense form to allow continuous-valued operations in subsequent diffusion-based and neural network modules:
\begin{equation}
\mathbf{X}_d = \mathrm{dense}(\mathbf{X}_{\mathrm{HVG}}),
\label{eq:dense}
\end{equation}

where $\mathbf{X}_d \in \mathbb{R}^{n \times k}$ is the fully realized dense gene expression matrix. 

\subsection{\mymethod{}}

Figure \ref{fig: pipeline} represents the whole pipeline of the proposed model, \mymethod{}. The overall method can be divided into three segments: MAGIC-based imputation, Spatial Attention, and Final Imputation Strategy.

\subsubsection{MAGIC-based Graph Diffusion}

To address sparsity in ST data due to dropout events, a manifold learning method using the Markov Affinity-based Graph Imputation of Cells (MAGIC)\cite{dijk2017magic} was used as depicted in Figure~\ref{fig: pipeline}. The model imputes missing values by representing expression data as a diffusion process on a graph constructed from local neighborhood relations among spots. To reduce the noise and improve the neighborhood graph robustness, gene expression matrix $\mathbf{X}_d \in \mathbb{R}^{n \times k}$ was optionally reduced by Principal Component Analysis (PCA)\cite{abdi2010principal}. Projection onto the top $d$ principal components results in:
\begin{equation}
\mathbf{Z} = \mathbf{X}_d \mathbf{W}_{\text{PCA}},
\label{eq:pca}
\end{equation}

where $\mathbf{W}_{\text{PCA}} \in \mathbb{R}^{k \times d}$ is the PCA loading matrix and $d = 100$ in the default configuration. PCA was bypassed when the original dimensionality was already sufficiently small, and $\mathbf{Z} = \mathbf{X}_d$ was used directly. 

After PCA,  a $k$-nearest neighbor (kNN) graph is formulated for local attention among the spots. For each spot, $\mathbf{z}_i \in \mathbb{R}^{d}$, the $k$ most similar samples were identified based on the Euclidean distance:
\begin{equation}
\mathcal{N}_i = \arg\min_{j \ne i} \| \mathbf{z}_i - \mathbf{z}_j \|_2, \quad |\mathcal{N}_i| = k,
\label{eq:knn}
\end{equation}

\noindent To prevent excessive connectivity in high-density areas, the number of neighbors was limited to a parameter $k_{\text{max}} = 3k$, where the value of k was set to 5. This cap ensures that each spot maintains a maximum of 15 neighbors, effectively balancing the capture of local biological relationships with the need to prevent over-smoothing during the diffusion process. 

From the kNN graph, an affinity matrix is derived using an adaptive Gaussian kernel. An affinity matrix is a weighted adjacency matrix that stores pairwise similarity scores between samples, and it is useful because it converts neighbor relationships into continuous edge weights that can be used by graph-based algorithm. Pairwise similarities were computed using an adaptive Gaussian kernel applied to the kNN distances. For each cell $i$, the local bandwidth $\sigma_i$ was defined as the median distance to its $k$ nearest neighbors, allowing the similarity scale to adjust to local data density. The affinity between spot $i$ and neighbor $j$ was computed as:
\begin{equation}
A_{ij} = \exp\left(-\frac{\|\mathbf{z}_i - \mathbf{z}_j\|_2^2}{2 \sigma_i^2} \right)^\alpha,
\label{eq:affinity}
\end{equation}

where $\alpha$ is a decay parameter controlling the sharpness of similarity weighting (default $\alpha=1$). Larger $\alpha$ values make similarities decrease more rapidly with distance, emphasizing very close neighbors and reducing the influence of farther ones. The resulting affinity matrix $\mathbf{A} \in \mathbb{R}^{n \times n}$ was symmetrized as:
\begin{equation}
\mathbf{W} = \frac{1}{2}(\mathbf{A} + \mathbf{A}^\top),
\label{eq:symmetric_affinity}
\end{equation}

Symmetry is required because we assumed undirected graphs, where edge weights must be equal in both directions. This operation ensures mutual consistency in neighborhood relationships by making the similarity between spots $i$ and $j$ identical regardless of direction. The symmetric affinity matrix $\mathbf{W}$ captures local similarity relationships between spots and therefore serves as the foundation for information propagation. To spread information across this graph while respecting its structure, a diffusion operator is applied. This is done by row-normalizing the affinity matrix $\mathbf{W}$ to obtain a stochastic transition matrix:
\begin{equation}
\mathbf{P} = \mathbf{D}^{-1} \mathbf{W}, \quad \text{where } D_{ii} = \sum_j W_{ij},
\label{eq:transition}
\end{equation}

where each row of $\mathbf{P}$ sums to 1 and represents transition probabilities from one spot to its neighbors. Diffusion is then performed by raising $\mathbf{P}$ to the $t$-th power:
\begin{equation}
\mathbf{P}^{(t)} = \mathbf{P}^t,
\label{eq:diffusion_power}
\end{equation}

which simulates a $t$-step random walk on the graph and allows information to propagate beyond immediate neighbors. The imputed expression matrix is then obtained as:
\begin{equation}
\mathbf{X}_{\text{MAGIC}} = \mathbf{P}^{t} \mathbf{X}_d,
\label{eq:diffusion}
\end{equation}

which smooths the observed expression values according to the graph structure. This diffusion process reduces noise and recovers dropout-affected gene expression while preserving the local biological neighborhood relationships encoded in the graph. $\mathbf{X}_{\text{MAGIC}}$ is used in the subsequent stages for final imputation.

\subsubsection{Spatial Transformer Attention}

A transformer encoder based on attention was used for the two-dimensional spatial coordinates to include spatial topological data in the imputation process. This allowed the model to acquire geometric representations that contain both local and global spatial relations in the architecture of the tissue. Spatial coordinates for all spots were used to form a 2D matrix $\mathbf{S} \in \mathbb{R}^{n \times 2}$, where each row $\mathbf{s}_i = (x_i, y_i)$ represents the physical location of $ith$ spots within the tissue section. These spatial properties were later encoded via a learned attention mechanism, enabling the model to learn spatial relationships and preserve tissue topology in imputation. Integration of $\mathbf{S}$ into the representation learning ensures that imputation is not only gene expression similarity-driven but also driven by spatial proximity and structural context.

Each spatial coordinate is embedded into a higher-dimensional representation through a linear transformation:
\begin{equation}
\mathbf{h}_i = \mathbf{W}_e \mathbf{s}_i + \mathbf{b}_e, \quad \mathbf{h}_i \in \mathbb{R}^{d_s},
\label{eq:spatial_embedding}
\end{equation}
where $d_s = 32$ denotes the spatial embedding dimension. This transformation maps the 2D spatial coordinates into a learnable feature space that can be processed jointly with other model components. The resulting spatial embedding matrix $\mathbf{H} \in \mathbb{R}^{n \times d_s}$ is then passed to a single-layer transformer encoder with a multi-head self-attention mechanism ($h=2$ heads). The encoder produces attention-weighted combinations of all spatial locations, yielding context-aware spatial representations:
\begin{equation}
\mathbf{H}_{\text{attn}} = \operatorname{TransformerEncoder}(\mathbf{H}),
\label{eq:transformer}
\end{equation}

Through self-attention, each spatial location interacts with every other location, allowing the model to learn both local and long-range spatial dependencies directly from the data. Because spatial coordinates are encoded into the embeddings, attention weights depend on relative spatial positions, enabling the model to capture geometric layouts without requiring explicitly defined adjacency matrices or manually chosen distance thresholds. This design provides a flexible way to represent complex spatial patterns by learning relationships between regions dynamically. In the next stage, the learned spatial features are projected to match the dimensionality of the gene expression space using a fully connected layer:
\begin{equation}
\mathbf{H}_{\text{proj}} = \mathbf{H}_{\text{attn}} \mathbf{W}_p + \mathbf{b}_p, \quad \mathbf{H}_{\text{proj}} \in \mathbb{R}^{n \times G},
\label{eq:spatial_projection}
\end{equation}
where the projection aligns spatial representations with the gene feature dimension. The projected spatial embeddings are then concatenated with the MAGIC-inspired-imputed expression matrix, $\mathbf{x}_{\text{MAGIC}} \in \mathbb{R}^{G}$ to construct a fused representation:
\begin{equation}
\mathbf{X}_{\text{fused}} = [\mathbf{X}_{\text{MAGIC}} \, \| \, \mathbf{H}_{\text{proj}}], \in \mathbb{R}^{2G}
\label{eq:fused_matrix}
\end{equation}
where $\|$ represents the concatenation of vectors. The final fused representations serve as the input to the subsequent neural fusion model. This fused matrix jointly incorporates molecular expression information and learned spatial context, allowing downstream modeling to utilize both gene-level similarity and spatial organization, particularly in regions where expression measurements are sparse or noisy.

\subsubsection{Final Imputation using Autoencoder Architecture}

A fully connected encoder--decoder network was employed to learn a nonlinear mapping from the fused input features to an enhanced gene expression representation. The encoder compresses the $2G$-dimensional fused input into a low-dimensional latent space through two successive transformations:
\begin{equation}
\mathbf{h}_1 = \mathrm{ReLU}(\mathbf{X}_{\text{fused}} \mathbf{W}_1 + \mathbf{b}_1), 
\quad \mathbf{h}_1 \in \mathbb{R}^{n \times 512},
\label{eq:encoder_layer1}
\end{equation}

\begin{equation}
\mathbf{h}_2 = \mathrm{ReLU}(\mathrm{Dropout}(\mathbf{h}_1)\mathbf{W}_2 + \mathbf{b}_2), 
\quad \mathbf{h}_2 \in \mathbb{R}^{n \times 256},
\label{eq:encoder_layer2}
\end{equation}

where ReLU activations introduce nonlinearity and help maintain non-negative reconstructed expression values.  

The decoder reconstructs gene expression from the latent representation by progressively mapping it back to the original gene dimension:
\begin{equation}
\hat{\mathbf{X}} = 
\mathrm{ReLU}\!\left(
\mathrm{ReLU}(\mathbf{h}_2 \mathbf{W}_3 + \mathbf{b}_3)\mathbf{W}_4 + \mathbf{b}_4
\right),
\quad \hat{\mathbf{X}} \in \mathbb{R}^{n \times G},
\label{eq:decoder}
\end{equation}

where $\hat{\mathbf{X}}$ denotes the reconstructed gene expression matrix. Model parameters were optimized by minimizing the mean squared reconstruction error:
\begin{equation}
\mathcal{L} =
\frac{1}{n}\sum_{i=1}^{n}
\left\|
\hat{\mathbf{x}}_i - \mathbf{x}_{\text{MAGIC},i}
\right\|_2^2,
\label{eq:loss}
\end{equation}
where $\hat{\mathbf{x}}_i$ represents the reconstructed expression profile of sample $i$.

To improve robustness and encourage recovery of missing values, a masking strategy was applied during training. A binary mask $\mathbf{M} \in \{0,1\}^{n \times G}$ was generated such that approximately $20\%$ of the MAGIC-imputed expression values were randomly set to zero:
\begin{equation}
\tilde{\mathbf{X}}_{\text{MAGIC}} =
\mathbf{X}_{\text{MAGIC}} \odot \mathbf{M},
\label{eq:dropout_mask}
\end{equation}
where $\odot$ denotes element-wise multiplication. The masked expression matrix was concatenated with the projected spatial embeddings to form corrupted input samples:
\begin{equation}
\tilde{\mathbf{X}}_{\text{fused}} =
\left[
\tilde{\mathbf{X}}_{\text{MAGIC}} \,\|\, \mathbf{H}_{\text{proj}}
\right],
\label{eq:dropout_fused}
\end{equation}

The network was trained to reconstruct the original (unmasked) $\mathbf{X}_{\text{MAGIC}}$ from $\tilde{\mathbf{X}}_{\text{fused}}$, enabling the model to infer missing gene expression values using both molecular expression context and learned spatial information. Optimization was performed using the Adam optimizer with learning rate $\eta$, trained for multiple epochs using mini-batch gradient descent with batch size $B = 256$. This final stage refines the initial graph-based imputation by learning nonlinear corrections guided jointly by spatial structure and expression similarity, improving reconstruction in regions affected by sparsity or measurement noise.
The entire dropout imputation process, \mymethod, is outlined in Algorithm \ref{alg:\mymethod{}}.

\begin{algorithm}[H]
\caption{Spatially Informed Dropout Imputation via MAGIC and Transformer Fusion (\mymethod{})}
\label{alg:\mymethod{}}
\begin{algorithmic}[1]
\REQUIRE Gene expression matrix $X \in \mathbb{R}^{N \times G}$; Spatial coordinates $C \in \mathbb{R}^{N \times 2}$; Parameters: $k$ (neighbors), $t$ (diffusion steps), $d$ (PCA dimensions), $e$ (embedding dimensions)
\ENSURE Final imputed matrix $\hat{X} \in \mathbb{R}^{N \times G}$

\STATE \textbf{Preprocessing:}
\STATE Select top $G'$ highly variable genes from $X$
\STATE Convert sparse $X$ to dense if needed
\STATE Update $X \gets X[:, 1{:}G']$, set $G \gets G'$

\STATE \textbf{MAGIC Graph Diffusion:}
\IF{$d < G$}
    \STATE Perform PCA: $X_{\mathrm{pca}} \gets \mathrm{PCA}(X, d)$
\ELSE
    \STATE $X_{\mathrm{pca}} \gets X$
\ENDIF
\STATE Construct $k$-NN graph from $X_{\mathrm{pca}}$
\STATE Compute affinity matrix $W$ using adaptive Gaussian kernel
\STATE Symmetrize: $W \gets \frac{1}{2}(W + W^{T})$
\STATE Row-normalize: $P_{ij} \gets \frac{W_{ij}}{\sum_j W_{ij}}$
\STATE Apply diffusion: $X_{\mathrm{magic}} \gets P^t \cdot X$

\STATE \textbf{Spatial Attention Embedding:}
\STATE Apply linear embedding: $E_C \gets \mathrm{Linear}(C)$
\STATE Pass through transformer encoder: $H_C \gets \mathrm{Transformer}(E_C)$
\STATE Project to gene space: $S \gets \mathrm{Linear}(H_C)$

\STATE \textbf{Fusion and Neural Imputation:}
\STATE Form fused input: $Z \gets [X_{\mathrm{magic}} \, || \, S]$
\FOR{each training batch $(Z_b, X_b)$}
    \STATE Sample dropout mask $M_b \sim \mathrm{Bernoulli}(p)$
    \STATE Apply dropout: $\tilde{X}_b \gets M_b \odot X_b$
    \STATE Form input: $\tilde{Z}_b \gets [\tilde{X}_b \, || \, S_b]$
    \STATE Encode: $h_b \gets f_{\mathrm{enc}}(\tilde{Z}_b)$
    \STATE Decode: $\hat{X}_b \gets f_{\mathrm{dec}}(h_b)$
    \STATE Compute loss: $\mathcal{L} \gets \mathrm{MSE}(\hat{X}_b, X_b)$
    \STATE Update network parameters
\ENDFOR

\STATE \textbf{Final Inference:}
\STATE Compute: $\hat{X} \gets f_{\mathrm{dec}}(f_{\mathrm{enc}}(Z))$

\end{algorithmic}
\end{algorithm}

\section{Results}\label{sec4}
In this section, we present a comprehensive evaluation of our proposed \mymethod{} framework across seven spatial transcriptomics datasets, highlighting consistent improvements in clustering accuracy over baseline methods. We further analyze differential gene expression patterns, demonstrating enhanced detection of biologically relevant signals, and perform pathway enrichment analysis to reveal meaningful metabolic and neural processes captured by overlapping genes. Together, these findings underscore the robustness, generalizability, and biological interpretability of \mymethod{} in spatial transcriptomics analysis.

\subsection{Clustering Performance Across Datasets}


To evaluate the effectiveness of our proposed framework, we evaluated it across a total of \textbf{seven spatial transcriptomics datasets}, including \textbf{three \stereoseq{}}, \textbf{three \slideseq{}}, and \textbf{one \scispace{}} dataset, shown in table \ref{tab:ari_scores}. The Adjusted Rand Index (ARI) \cite{warrens2022understanding} was used as the evaluation metric to quantify clustering accuracy relative to ground truth cell-type annotations, comparing the raw data before imputation against four distinct computational strategies. These include MAGIC, a standard Markov affinity-based graph diffusion method, and two hybrid baselines: Attention PCA and Attention UMAP. Attention PCA integrates the context-aware spatial representations generated by the Spatial Transformer Attention module, which maps 2D coordinates into a 32-dimensional learned feature space to capture both local and global geometric dependencies, with gene expression data using Principal Component Analysis (PCA) for linear dimensionality reduction \cite{greenacre2022principal}. Similarly, Attention UMAP leverages these transformer-derived spatial embeddings but employs Uniform Manifold Approximation and Projection (UMAP) to preserve complex, non-linear structural relationships within the tissue architecture \cite{mcinnes2018umap}. Finally, we evaluate SpatialMAGIC, our proposed framework that combines graph-based molecular diffusion with spatial self-attention features through a dedicated refinement module to achieve high precision in biological signal representation.

\begin{table*}[ht!]
\caption{Comparison of ARI scores across different spatial transcriptomics datasets and imputation methods. \mymethod{} consistently outperforms baselines across all dataset types.}
\label{tab:ari_scores}
\centering
\resizebox{\textwidth}{!}{%
\begin{tabular}{llcccccc}
\toprule
\textbf{Dataset Type} & \textbf{Dataset} & \textbf{Before Imputation} & \textbf{MAGIC} & \textbf{Attention PCA} & \textbf{Attention UMAP} & \textbf{\mymethod{}} \\
\midrule
\multirow{3}{*}{\stereoseq{}} 
  & DX6\_D2\_stereo-seq       & 0.2661 & 0.2889 & 0.2818 & 0.2839 & \textbf{0.3254} \\
  & DT2\_D0\_stereo-seq       & 0.2847 & 0.3088 & 0.3194 & 0.3014 & \textbf{0.3301} \\
  & FB2\_D1\_stereo-seq       & 0.1679 & 0.2192 & 0.2580 & 0.2543 & \textbf{0.2993} \\
\midrule
\multirow{3}{*}{\slideseq{}}
  & stickles\_mouse\_slideseq & 0.1740 & 0.1764 & \textbf{0.2249} & 0.1736 & 0.2193 \\
  & slide\_seq\_diabetes1\_T4 & 0.2657 & 0.2577 & 0.2336 & 0.2197 & \textbf{0.2688} \\
  & slide\_seq\_WT1\_T3       & 0.2891 & 0.3049 & 0.2870 & 0.2287 & \textbf{0.3074} \\
\midrule
SciSpace & GSE166692\_scispace & 0.3095 & 0.4020 & 0.3717 & 0.3316 & \textbf{0.4216} \\
\bottomrule
\end{tabular}%
}
\end{table*}

\begin{figure*}[ht!]
  \centering
  \includegraphics[width=1\textwidth,height=.5\textheight]{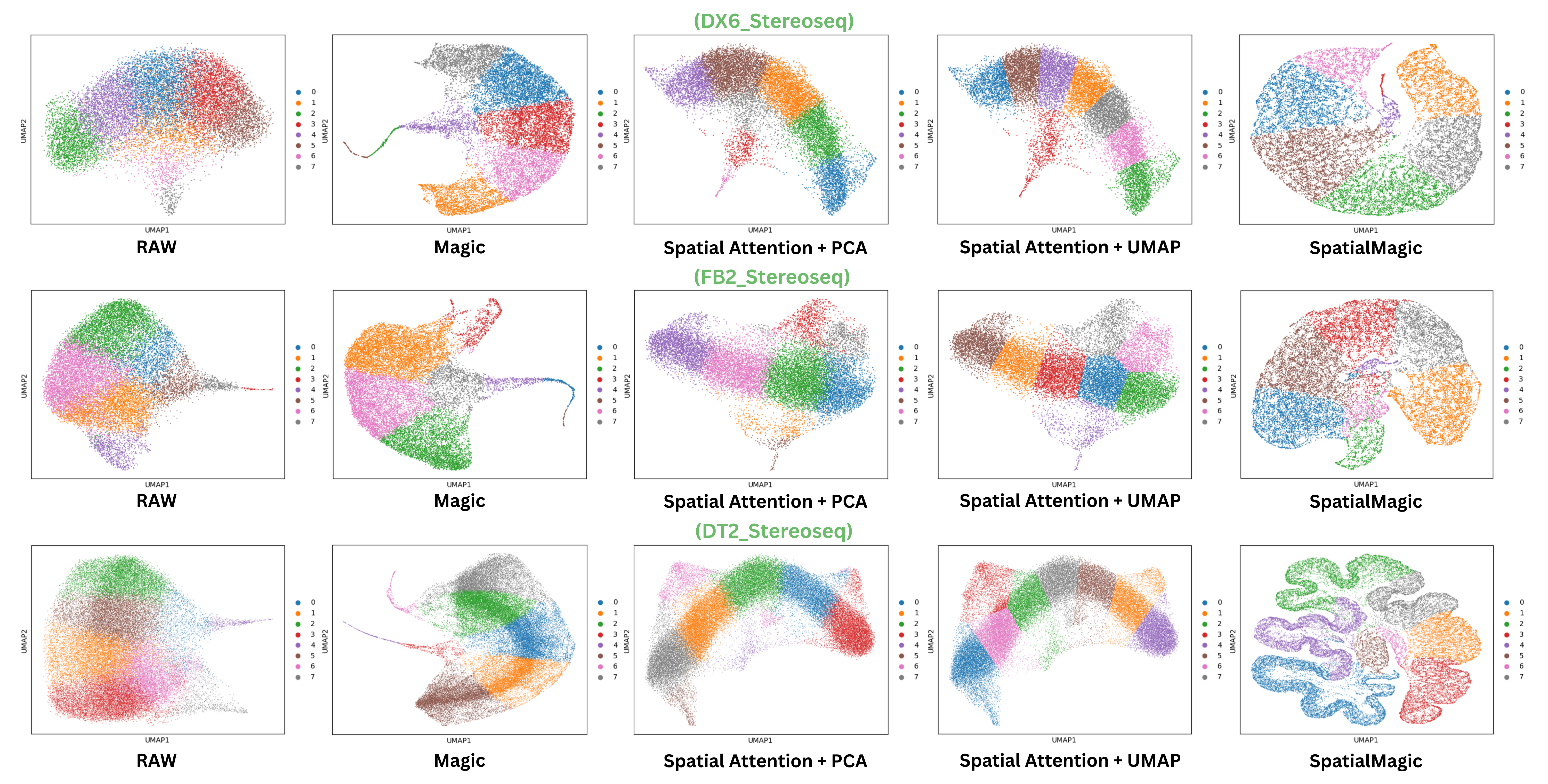}
  \caption{Clustering results of the \stereoseq{} dataset before and after imputations.}
 \label{fig:clust}
\end{figure*}

\textbf{\stereoseq{} datasets:} On the DX6\_D2\_stereo-seq dataset, our method improved the ARI from \textbf{0.2661 (raw)} to \textbf{0.3254}, outperforming both MAGIC (0.2889) and attention-based variants (Table~\ref{tab:ari_scores}). Similarly, on DT2\_D0\_stereo-seq, \mymethod{} achieved the highest ARI of \textbf{0.3301}, up from 0.2847 (raw) and exceeding the scores of MAGIC (0.3088), Attention PCA (0.3194), and Attention UMAP (0.3014). On FB2\_D1\_stereo-seq, the ARI increased from \textbf{0.1679} to \textbf{0.2993}, again surpassing MAGIC (0.2192) and all attention-based baselines. These results demonstrate the robustness of \mymethod{} in recovering biologically meaningful spatial patterns in high-resolution \stereoseq{} data, which is often sparse and noisy. Figure \ref{fig:clust} represents the visual clustering patterns of the \stereoseq{} dataset for the raw data and after applying dropout imputation methods, including \mymethod{}, enabling a direct comparison of their effects. 

\textbf{\slideseq{} datasets:} On the stickles\_mouse\_slideseq dataset, the proposed method performed comparably to the best attention variant (0.2193 vs.\ 0.2249 with Attention PCA), yet still outperformed the raw input (0.1740) and MAGIC (0.1764). For the slide\_seq\_diabetes1\_T4 dataset, \mymethod{} yielded the highest ARI of \textbf{0.2688}, improving upon raw (0.2657), MAGIC (0.2577), and attention variants (0.2336, 0.2197). On slide\_seq\_WT1\_T3, it achieved an ARI of \textbf{0.3074}, outperforming MAGIC (0.3049) and showing strong resilience to variation across methods. These results suggest that even in medium-resolution \slideseq{} data, \mymethod{} retains an edge in spatial domain recovery.

\textbf{\scispace{} dataset:} For the GSE166692\_scispace dataset, \mymethod{} significantly enhanced the ARI from \textbf{0.3095} to \textbf{0.4216}, outperforming MAGIC (0.4020), Attention PCA (0.3717), and Attention UMAP (0.3316). This substantial gain underlines the method’s ability to generalize across platforms and its effectiveness in leveraging spatial context to enhance gene expression imputation and clustering.

\begin{figure}[ht!]
  \centering
  \includegraphics[width=0.9\textwidth]{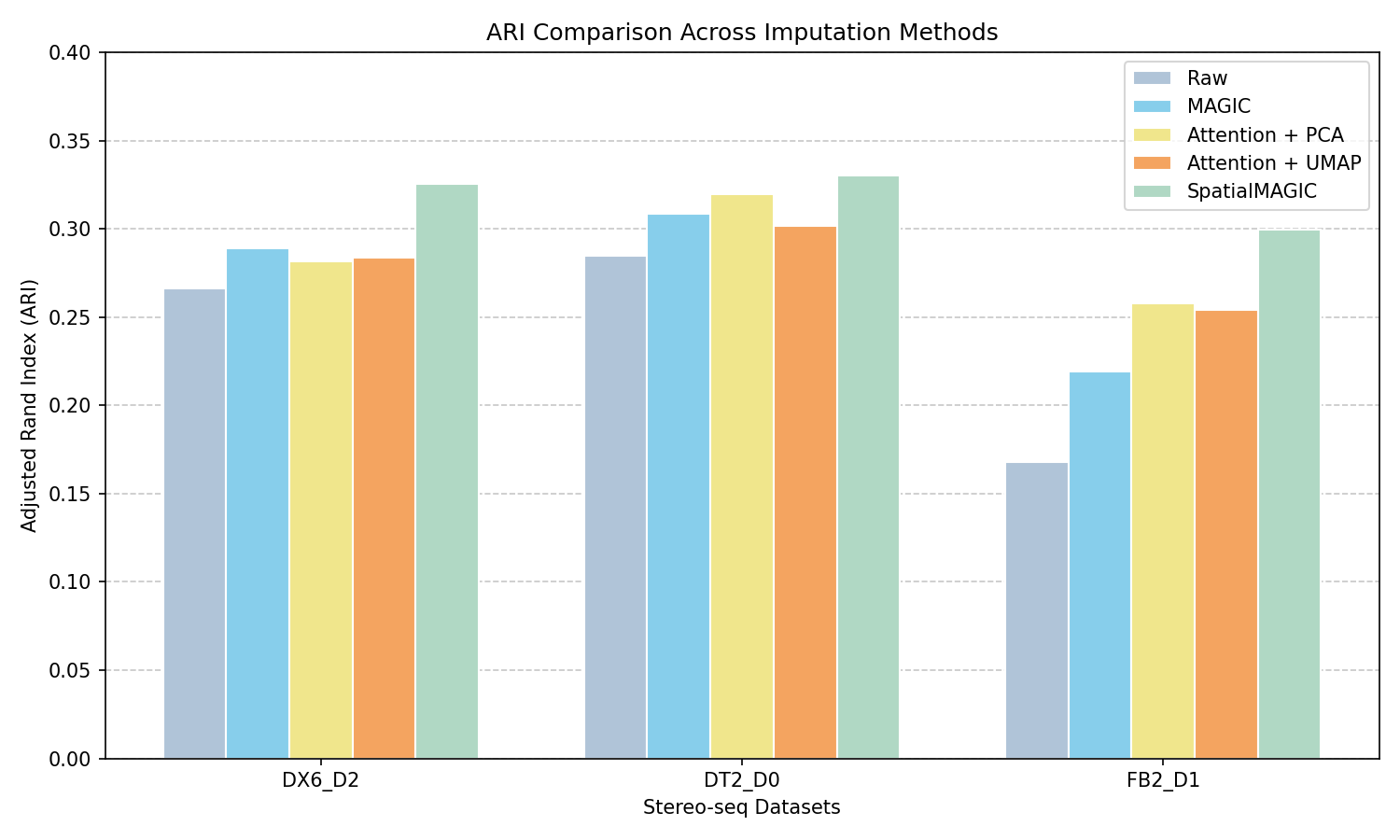}
  \caption{ARI Comparison Across Imputation Methods (\stereoseq{}).}
 \label{stereo_ari}
\end{figure}

Figure~\ref{stereo_ari} presents the Adjusted Rand Index (ARI) scores for three \stereoseq{} datasets (DX6\_D2, DT2\_D0, and FB2\_D1) across five imputation strategies: Raw (without imputation), MAGIC, Spatial Attention + PCA, Spatial Attention + UMAP, and the proposed \mymethod{}.


\begin{figure}[ht!]
\centering
\includegraphics[width=0.9\textwidth]{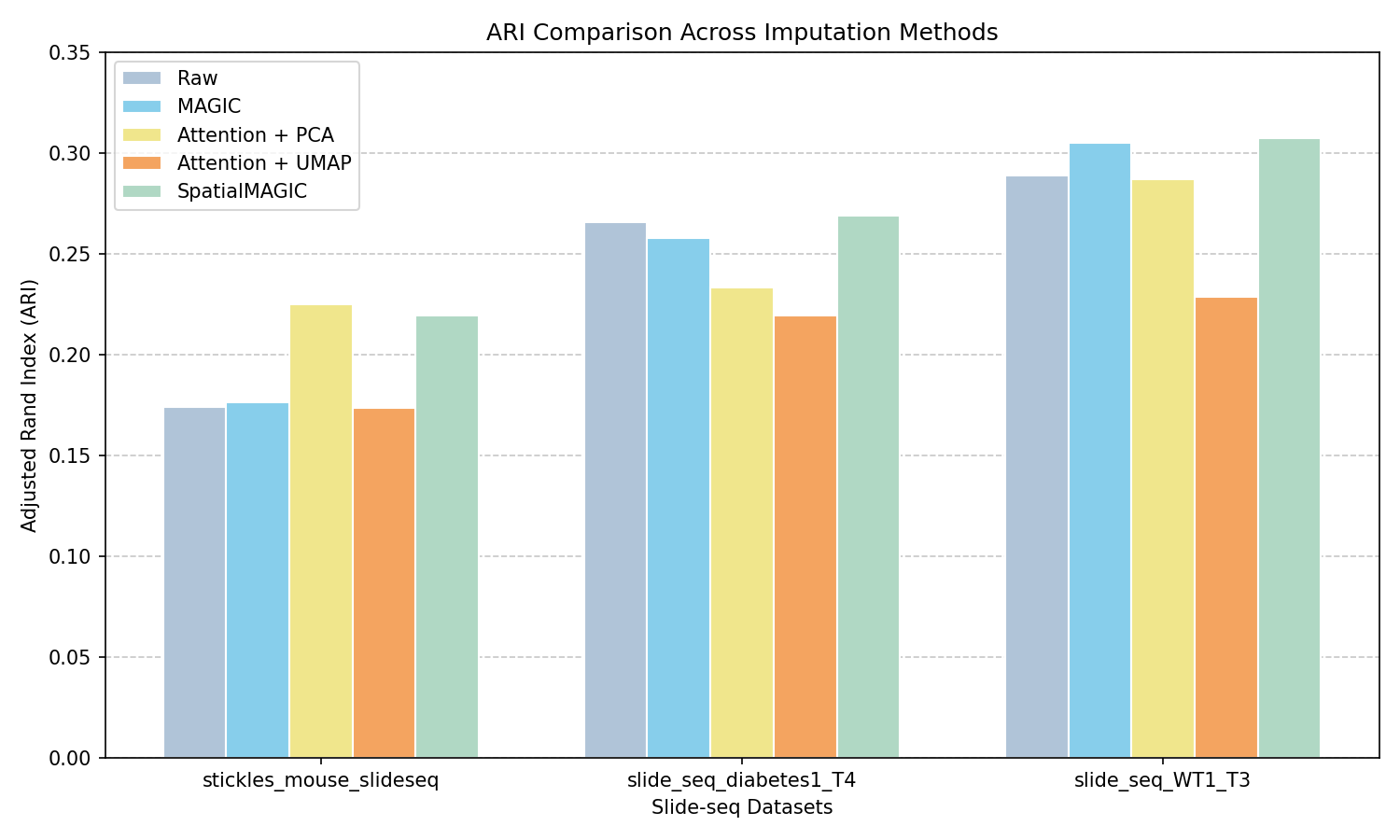}
\caption{ARI Comparison Across Imputation Methods (\slideseq{}).}
\label{slide_ari}
\end{figure}

\begin{figure}[!ht]
\centering
\includegraphics[width=0.9\textwidth]{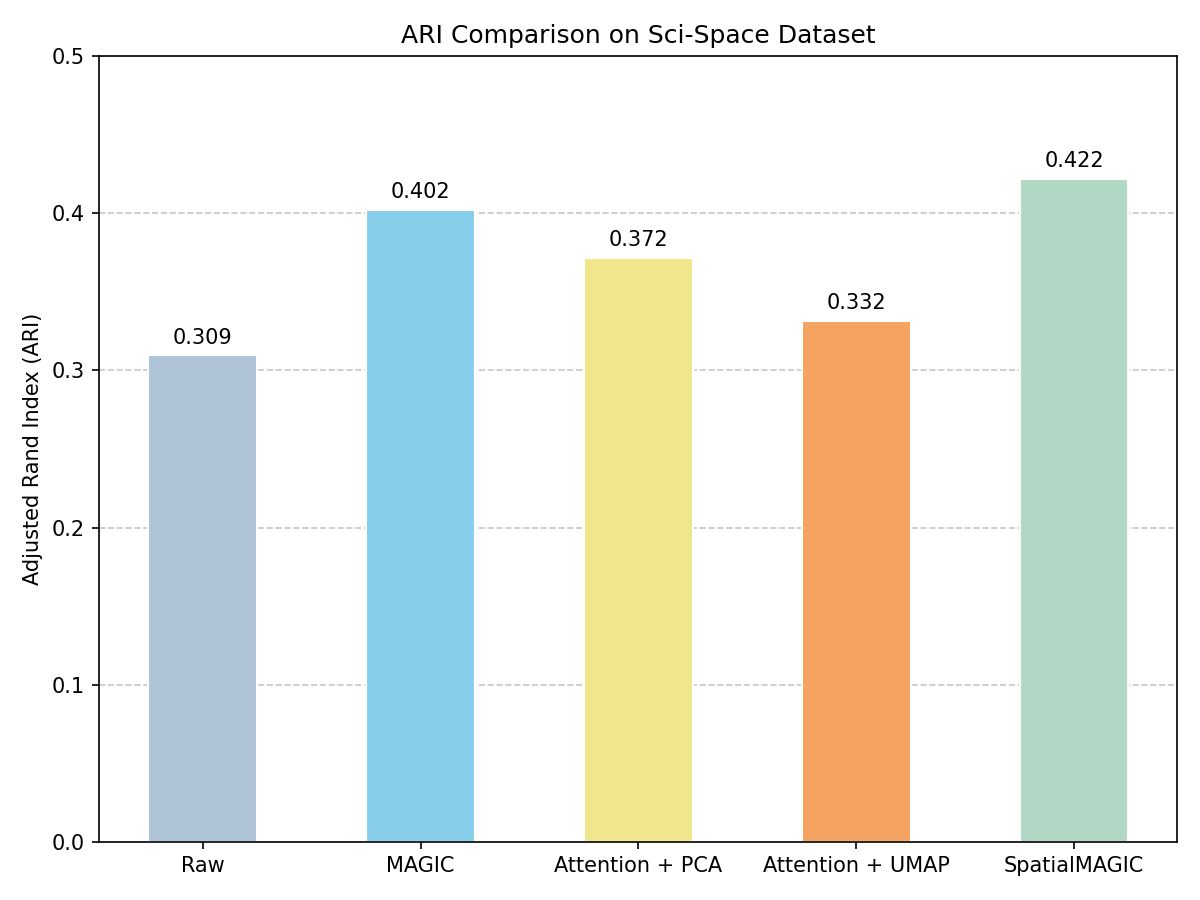}
\caption{ARI comparison across imputation methods (\scispace{}).}
\label{scispace_ari}
\end{figure}

\noindent Figure \ref{slide_ari} presents the Adjusted Rand Index (ARI) scores for three \slideseq{} datasets (stickles\_mouse\_slideseq, slide\_seq\_diabetes1\_T4, and slide\_seq\_WT1\_T3) across five imputation strategies: Raw (without imputation), MAGIC, Spatial Attention + PCA, Spatial Attention + UMAP, and the proposed \mymethod{}.
Figure \ref{scispace_ari} presents the Adjusted Rand Index (ARI) score for the \scispace{} dataset (GSE166692\_scispace) across the same five imputation methods: Raw (without imputation), MAGIC, Spatial Attention + PCA, Spatial Attention + UMAP, and \mymethod{}.

The runtime comparison between MAGIC and \mymethod{} across three benchmark datasets is presented in Table~\ref{tab:runtime}. \mymethod{} consistently exhibits higher computational overhead compared to MAGIC, which is attributable to the additional spatial processing it performs. Notably, the most significant performance gap is observed on the DT2 dataset, where \mymethod{} incurs a runtime of 1451.36 seconds compared to 334.75 seconds for MAGIC, corresponding to a slowdown of approximately 4.3× times. On the DX6 and FB2 datasets, the runtime inconsistency is comparatively modest, with \mymethod{} requiring 1.52× and 1.18× times of the runtime of MAGIC, respectively.

\begin{table}[h]
\centering
\begin{tabular}{lcc}
\hline
\textbf{Dataset} & \textbf{MAGIC (s)} & \textbf{\mymethod{} (s)} \\
\hline
DX6 & 191.50  & 291.56   \\
DT2 & 334.75  & 1451.36  \\
FB2 & 282.91  & 332.74   \\
\hline
\end{tabular}
\caption{Runtime comparison of MAGIC and \mymethod{} across  \stereoseq{} datasets. All experiments were conducted using the cloud-based notebook environment provided by Kaggle. The implementation was executed on a GPU-accelerated runtime equipped with two NVIDIA Tesla T4 GPUs, each with 15 GiB of VRAM, and approximately 30 GiB of system RAM.}
\label{tab:runtime}
\end{table}

\subsection{Differential Expression and Overlapping Gene Analysis Before and After \mymethod{}}
To assess the effect of the proposed \mymethod{} model on differential gene expression patterns, a comparative analysis was conducted between pre- and post-imputation data sets of all three \stereoseq{} samples. Initially, highly variable genes were identified using DESeq2, for which a threshold of a p-value of less than 0.01 was considered for significant relevance. Genes were then categorized as being up-regulated, down-regulated, and not significant according to log2 fold-change and adjusted p-values. This analysis was performed independently for raw expression matrices and for matrices imputed by \mymethod{} to study how a model affects gene-level regulation.

\begin{figure*}[ht!]
  \centering
  \includegraphics[width=0.99\textwidth]{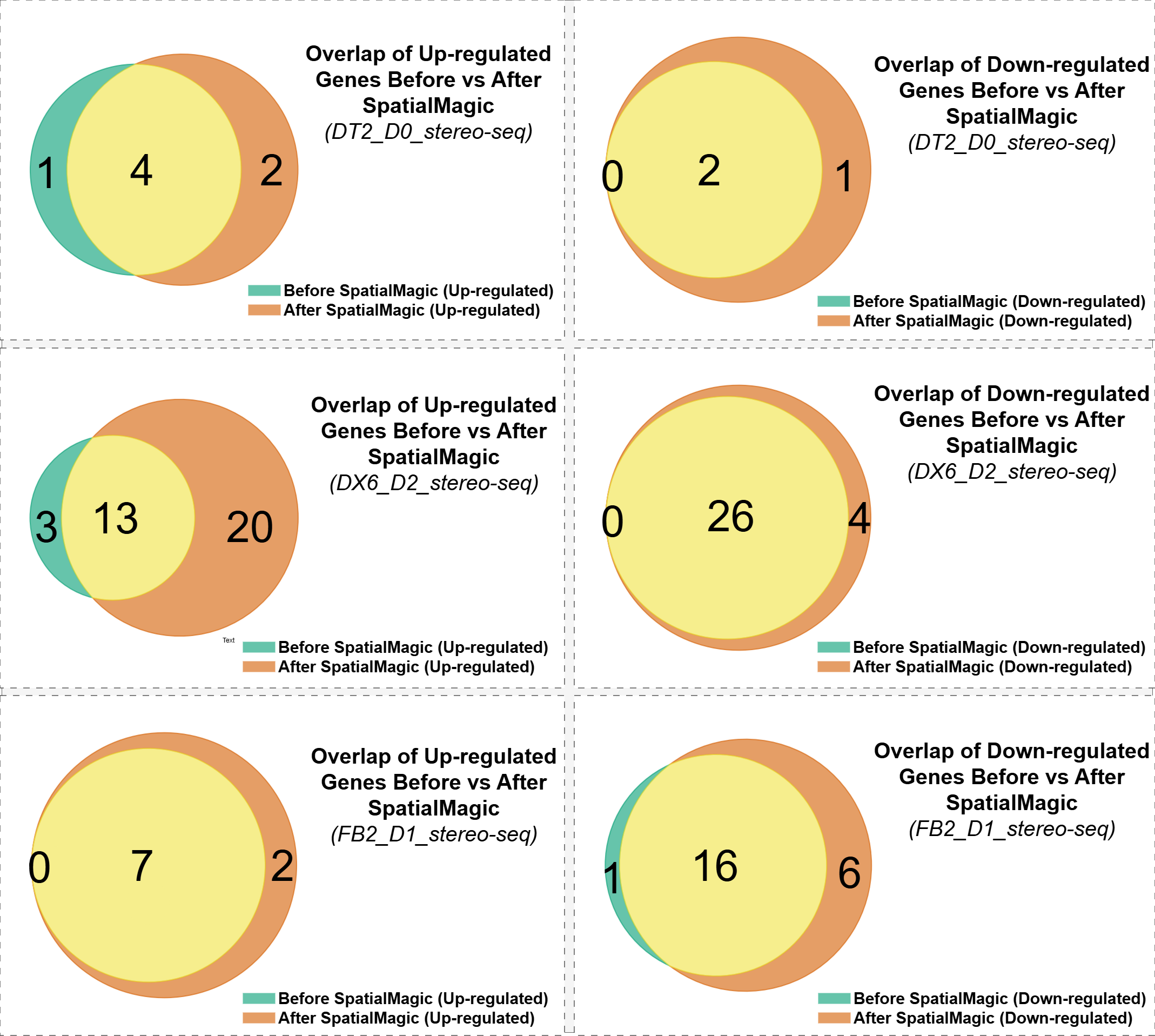}
  \caption{Overlapping up- and down-regulated genes before and after applying \mymethod{} across the three \stereoseq{} datasets (DT2\_D0, DX6\_D2, and FB2\_D1). Each pair of Venn diagrams represents the overlap and enhancement of significant gene detection after imputation.}
 \label{fig:venn_diagram}
\end{figure*}

For each dataset, two Venn diagrams were constructed, one for up-regulated and one for down-regulated genes to illustrate the intersection of gene sets before and after applying \mymethod{} (Fig. \ref{fig:venn_diagram}). The overlapping regions indicate genes exhibiting consistent regulatory patterns, whereas the non-overlapping areas represent genes newly detected or corrected following the imputation process. In the DT2\_D0 dataset, 1 and 2 genes were identified as up-regulated before and after \mymethod{}, respectively, having an overlap of 4 genes. Similarly, 0 and 1 genes were down-regulated before and after imputation, with 2 genes overlapping. For the DX6\_D2 dataset, 3 and 20 genes were found to be up-regulated before and after \mymethod{}, respectively, with 13 overlapping genes, while 0 and 5 genes were down-regulated, having an overlap of 26 genes. In the FB2\_D1 dataset, 0 and 2 genes were up-regulated before and after imputation, respectively, with 7 overlapping genes, whereas 1 and 6 genes were down-regulated, with 16 overlapping genes.

SpatialMAGIC significantly enhances the detection of biologically informative gene expression signals that are often obscured by technical noise and sparsity in raw spatial transcriptomics data. The substantial overlaps across datasets further confirm that \mymethod{} preserves intrinsic regulatory patterns while enhancing the detection of differentially expressed genes. The Venn diagrams summarized in Figure \ref{fig:venn_diagram} illustrate the relationships, supporting that the model improves analysis of differential expression without compromising biological relevance. Table \ref{tab:gene_list} shows the detailed list of up- and down-regulated genes across the \stereoseq{} datasets. It includes the genes identified before imputation, the genes obtained after applying \mymethod{}, and the overlapping genes shared between the two sets. Our method reveals several biologically relevant genes that were not detected prior to imputation across the \stereoseq{} datasets. For example, in the DT2\_D0 dataset, \mymethod{} identified Ephx2 and Nop9 as up-regulated and Dnajb11 as down-regulated, which were not captured in the raw data analysis. Notably, Ephx2 encodes soluble epoxide hydrolase, an enzyme involved in lipid metabolism and inflammatory regulation that has been linked to cardiovascular and metabolic diseases \cite{luria2007compensatory}. In the DX6\_D2 dataset, the method uncovered 24 additional genes, including up-regulated genes like Mdm2, Ap2a2, and Plg, alongside down-regulated genes such as Slc4a4 and Rdh11. Among these, Mdm2 is a well-established regulator of the p53 tumor suppressor pathway and plays a critical role in cell cycle control and tumorigenesis \cite{nag2013mdm2}, while Plg, encoding plasminogen, is essential for fibrinolysis and extracellular matrix remodeling during tissue repair \cite{ny2020plasminogen}. Furthermore, in the FB2\_D1 dataset, \mymethod{} detected genes such as Abhd17a and Asgr1, which are associated with protein depalmitoylation signaling processes and hepatic glycoprotein clearance respectively \cite{shi2025unconventional, svecla2024asgr1}. By recovering these biologically significant genes that were previously undetected, \mymethod{} provides a more comprehensive and biologically meaningful representation of the spatial transcriptomic landscape, thereby facilitating improved interpretation of tissue organization and molecular activity.


\begin{table}[h!]
\centering
\caption{Comparison of differentially expressed genes (DEGs) identified in \stereoseq{} datasets before and after \mymethod{} imputation.}
\label{tab:gene_list}
\small
\renewcommand{\arraystretch}{1.0}
\begin{tabular}{l l p{7cm}}
\toprule
\textbf{Dataset} & \textbf{Category} & \textbf{Genes} \\
\midrule

\multicolumn{3}{l}{\textbf{Up-Regulated}} \\
\midrule

\multirow{3}{*}{DT2}
  & Overlapping  & \textit{\footnotesize Glul, Cyp2c50, Slco1b2, Rnase4} \\
  & Before Applying Algorithm  & \textit{\footnotesize Mup18} \\
  & After SpatialMAGIC & \textit{\footnotesize Ephx2, Nop9} \\
\cmidrule{1-3}

\multirow{3}{*}{DX6}
  & Overlapping  & \textit{\footnotesize Hal, Gls2, Sfxn1, Hsd17b13, Apoc2, Mt1, Pabpc1, Ube2c, Sult1a1, Cps1, Selenbp2, Fh1, Alb} \\
  & Before Applying Algorithm  & \textit{\footnotesize Tpi1, Btg2, Hpx} \\
  & After SpatialMAGIC & \textit{\footnotesize Ap2a2, Eif3d, Ak3, Uba5, Dnajc22, Sacm1l, Tor1a, Ide, Etfb, Crk, Mkrn2, Hs3st3b1, Mdm2, Eif6, Vps35, Zc3h15, Nr1h3, Nhp2, Acads, Plg} \\
\cmidrule{1-3}

\multirow{3}{*}{FB2}
  & Overlapping  & \textit{\footnotesize Cib3, C9, Spp1, Arg1, Selenop, Trf, Sds} \\
  & Before Applying Algorithm  & \textit{\footnotesize ---} \\
  & After SpatialMAGIC & \textit{\footnotesize Abhd17a, Fam47e} \\
\midrule

\multicolumn{3}{l}{\textbf{Down-Regulated}} \\
\midrule

\multirow{2}{*}{DT2}
  & Overlapping  & \textit{\footnotesize Hsd17b13, Alb} \\
  & Before Applying Algorithm  & \textit{\footnotesize ---} \\
  & After SpatialMAGIC & \textit{\footnotesize Dnajb11} \\
\cmidrule{1-3}

\multirow{3}{*}{DX6}
  & Overlapping  & \textit{\footnotesize Glul, Insig1, Cyp4f15, Slc1a2, Cyp2d40, Mup18, Hmgcs1, B3gnt8, Slc10a1, Slco1b2, Ank2, Cyp2c50, Msrb1, Pon1, Mup17, Cyp2e1, Gsta3, Oat, Slc13a3, Cyp2c37, Acss2, Gulo, Mup11, Cyp4a14, Aldh1a7, Pah} \\
  & Before Applying Algorithm  & \textit{\footnotesize ---} \\
  & After SpatialMAGIC & \textit{\footnotesize Slc4a4, Dnaja3, Gtf2h5, Rdh11} \\
\cmidrule{1-3}

\multirow{3}{*}{FB2}
  & Overlapping  & \textit{\footnotesize Slc1a2, Cyp2c67, Glul, Gsta3, Mup18, Cyp1a2, Mup17, Cyp2e1, Oat, Pon1, Mup9, Mup16, Ang, Aldh3a2, Lect2, Lcat} \\
  & Before Applying Algorithm  & \textit{\footnotesize Rgn} \\
  & After SpatialMAGIC & \textit{\footnotesize Nudc, Asgr1, Laptm4b, Stx5a, Ly6d, Me1} \\
\bottomrule
\end{tabular}
\end{table}

\subsection{Pathway Analysis}

To investigate the biological functions represented by the shared genes across different \stereoseq{} datasets, we performed pathway enrichment analysis using the Reactome database \cite{milacic2024reactome} through the DAVID functional annotation tool \cite{sherman2022david,huang2009systematic}. Two pairwise comparisons were considered between DT2-DX6 and DX6-FB2. The results are summarized in Tables~\ref{tab:pathway_dt2_dx6} and \ref{tab:pathway_dx6_fb2}.

For the DT2-DX6 comparison, several pathways related to core metabolic activities and molecular transport were enriched. Recycling of bile acids and salts (R-MMU-159418) and Heme degradation (R-MMU-189483) were the most prominent, with high fold enrichment values of 272.85 and 244.13, respectively. Both pathways involved SLCO1B2 and ALB, which are known to play essential roles in transport and metabolic regulation \cite{slijepcevic2017hepatic}. Additional enrichment was observed in pathways associated with general metabolism, lipid metabolism, and steroid metabolism, reflecting the active biochemical environment during tissue development.

\begin{table*}[ht!]
\centering
\caption{Reactome pathway enrichment analysis of overlapping genes between DT2 and DX6 \stereoseq{} datasets.}
\label{tab:pathway_dt2_dx6}
\scriptsize
\renewcommand{\arraystretch}{1.3}
\begin{tabular}{p{4.5cm}p{1.2cm}p{3.5cm}p{1.5cm}p{1cm}}
\toprule
Term & P-Value & Genes & Fold Enrichment \\
\midrule
R-MMU-159418~Recycling of bile acids and salts & 0.005488 & SLCO1B2, ALB & 272.85 \\
R-MMU-189483~Heme degradation & 0.006132 & SLCO1B2, ALB & 244.13 \\
R-MMU-1430728~Metabolism & 0.007920 & SLCO1B2, ALB, HSD17B13, GLUL & 5.01 \\
R-MMU-189445~Metabolism of porphyrins & 0.010314 & SLCO1B2, ALB & 144.95 \\
R-MMU-556833~Metabolism of lipids & 0.013069 & SLCO1B2, ALB, HSD17B13 & 11.10 \\
R-MMU-194068~Bile acid and bile salt metabolism & 0.013842 & SLCO1B2, ALB & 107.87 \\
R-MMU-9748784~Drug ADME & 0.036734 & SLCO1B2, ALB & 40.33 \\
R-MMU-8957322~Metabolism of steroids & 0.040514 & SLCO1B2, ALB & 36.52 \\
\bottomrule
\end{tabular}
\end{table*}

Notably for the DX6–FB2 comparison, Astrocytic Glutamate-Glutamine Uptake and Metabolism (R-MMU-210455) and Neurotransmitter Uptake and Metabolism in Glial Cells (R-MMU-112313) were highly enriched, both involving SLC1A2(Also known as GLT-1, EAAT2) and GLUL, with fold enrichment values exceeding 770. These pathways are critical for neurotransmitter recycling and metabolic support in brain tissues, suggesting that shared genes between DX6 and FB2 are functionally linked to neural signaling and metabolic regulation \cite{pajarillo2019role,takahashi2015glutamate,ben2021evidence}. Additional enrichment was found in Drug ADME (R-MMU-9748784), general metabolism (R-MMU-1430728), and Glutamate and Glutamine Metabolism (R-MMU-8964539), highlighting both metabolic and signaling-related processes.

\begin{table*}[ht!]
\centering
\caption{Reactome pathway enrichment analysis of overlapping genes between DX6 and FB2 \stereoseq{} datasets.}
\label{tab:pathway_dx6_fb2}
\scriptsize
\renewcommand{\arraystretch}{1.3}
\begin{tabular}{p{4.5cm}p{1.2cm}p{3.5cm}p{1.5cm}p{1cm}}
\toprule
Term & P-Value & Genes & Fold Enrichment \\
\midrule
R-MMU-9748784~Drug ADME & 0.001487 & GSTA3, PON1, CYP2E1 & 40.33 \\
R-MMU-210455~Astrocytic Glutamate-Glutamine Uptake And Metabolism & 0.002154 & SLC1A2, GLUL & 773.08 \\
R-MMU-112313~Neurotransmitter uptake and metabolism In glial cells & 0.002154 & SLC1A2, GLUL & 773.08\\
R-MMU-1430728~Metabolism & 0.006631 & OAT, GSTA3, PON1, CYP2E1, GLUL & 4.18 \\
R-MMU-8964539~Glutamate and glutamine metabolism & 0.006988 & OAT, GLUL & 237.87 \\
\bottomrule
\end{tabular}
\end{table*}

Overall, the pathway analysis across both comparisons highlights the involvement of shared genes in key metabolic, transport, and neural signaling processes. In the DT2–DX6 comparison, metabolic and bile acid pathways dominate, reflecting active tissue development and molecular transport functions. In contrast, the DX6–FB2 comparison reveals pathways related to neurotransmitter metabolism and astrocytic activity, which are central to brain tissue organization and functional maturation. These findings indicate that the overlapping genes capture biologically meaningful transitions in metabolic and neural processes across developmental stages and tissue types.

\section{Discussion and Conclusion}\label{sec5}
The proposed \mymethod{} is an imputation model capable of overcoming data sparseness and noise in spatial transcriptomics based on graph diffusion and spatial self-attention. After being tested on numerous datasets from \stereoseq{}, \slideseq{}, and \scispace{} platforms, the model outperformed existing imputation methods under all tested datasets based on clustering accuracy and biological interpretability. The model improved the ARI for all datasets with good spatial coherence and strong denoising. Beyond numerical performance metrics, the imputed spatial transcriptomics data by \mymethod{} preserved biologically informative expression patterns and enhanced the detection of genes that are up- as well as down-regulated, as validated by overlapping gene and pathway studies, including metabolic, neural, and transport processes.


The enhanced performance is the result of mutual interaction between the graph diffusion process and spatial attention mechanisms. The complex feature representations can be captured from gene expression data by the diffusion process through tissue-level neighborhoods, thereby reducing dropouts and retaining long-range dependencies. The spatial attention module learns the local spatial patterns, and this ensures the consistent tissue structure in the imputed data. The model is being fused by combining spatial and learned features from the gene expression data into a single representation. This fusion makes the model more efficient than standard graph-based diffusion methods such as MAGIC \cite{roopra2020magic}.

\mymethod{} is more versatile and flexible compared to the existing methods. While the model shows higher computational overhead compared to MAGIC, particularly on larger datasets like DT2, where it showed a slowdown of approximately 4.3×. It consistently provides superior biological interpretability and spatial coherence across various platforms. It can clearly distinguish between spatial locations and gene expression data and find out significant biologically relevant results. The model can enhance spatial clustering, determine tissue domains, and explore changes in gene expression. 


However, the transformer module in \mymethod{} can be computationally expensive, especially when working with very high-resolution datasets. Experimental validation with biological samples is also necessary to further confirm the model’s effectiveness. Future extensions could explore incorporating multimodal inputs such as histological or proteomic data, while improving efficiency through sparse attention mechanisms and enhancing interpretability with attention visualization tools. Overall, \mymethod{} provides a robust, scalable, and interpretable framework for spatial omics analysis, enabling the generation of biologically insightful gene expression landscapes.

In summary, this study introduced \mymethod{}, a hybrid computational framework designed to address the challenges of data sparsity and technical noise in ST by integrating graph diffusion and spatial self-attention mechanisms. By employing MAGIC-based diffusion to address the sparse and high-dimensional nature of ST data, together with the spatial attention to extract the spatially localized patterns, \mymethod{} effectively reconstructs the biologically meaningful maps of gene expression while preserving the underlying tissue architecture. The utilization of the fusion-based refinement strategy further enhances performance by transforming the expression and spatial embeddings into an equivalent unified latent representation. The overlapping gene as well as the differential expression study revealed the capability of the framework to recover the suppressed regulatory information, while the pathway enrichment study revealed the recovery of the biologically related processes associated with metabolism, molecular transport, and neural signaling. These outcomes collectively establish that the \mymethod{} enhances the biological interpretability alongside the quantitative reliability of the ST data.

While \mymethod{} exhibits notable performance on multiple datasets, there are still multiple domains where extensive analysis is required. The performance of the approach on larger-scale tissue benchmarking studies, together with additional ST modalities, would provide stronger evidence for scalability and generalizability. Future work will focus on the incorporation of multimodal data, such as the inclusion of histological imaging data and proteomic information, optimizing computational efficiency for large-scale inference, and developing interpretability tools for visualizing spatial attention maps.

\bibliographystyle{IEEEtran}
\bibliography{reference}

\end{document}